\documentclass[conference]{IEEEtran}
\IEEEoverridecommandlockouts
\usepackage{cite}
\usepackage{amsmath,amssymb,amsfonts}
\usepackage{algorithmic}
\usepackage{graphicx}
\usepackage{textcomp}
\usepackage{xcolor}
\usepackage{flushend}
\def\BibTeX{{\rm B\kern-.05em{\sc i\kern-.025em b}\kern-.08em
    T\kern-.1667em\lower.7ex\hbox{E}\kern-.125emX}}

\usepackage{caption}
\usepackage{subcaption}

\usepackage[ruled,vlined,linesnumbered]{algorithm2e}

\usepackage[left=1.62cm,right=1.62cm,top=1.9cm]{geometry}

\renewcommand{\baselinestretch}{1.01}

\begin{document}

\title{Towards Explainable Federated Learning: Understanding the Impact of Differential Privacy}

\author{\IEEEauthorblockN{Julio~Oliveira\IEEEauthorrefmark{1}, Rodrigo~Ferreira\IEEEauthorrefmark{1},
André~Riker\IEEEauthorrefmark{1}\IEEEauthorrefmark{2}, Glaucio~H.~S.~Carvalho\IEEEauthorrefmark{3}, Eirini~Eleni~Tsilopoulou\IEEEauthorrefmark{2}}
\IEEEauthorblockA{\IEEEauthorrefmark{1} Institute of Exact and Natural Sciences, Federal University of Par\'a, Brazil}
\IEEEauthorblockA{\IEEEauthorrefmark{2} Performance and Resource Optimization in Networks -- PROTON Lab\\ School of Electrical, Computer and Energy Engineering, Arizona State University, Tempe, AZ, USA}
\IEEEauthorblockA{\IEEEauthorrefmark{3}Faculty of Mathematics \& Science, Brock University, Canada}%
Email: \{julio.costa.oliveira, rodrigo.ferreira\}@icen.ufpa.br, gdecarvalho@brocku.ca, \{riker.a, eirini\}@asu.edu
\\[-3.0ex]
}
\maketitle

\begin{abstract}

Data privacy and eXplainable Artificial Intelligence (XAI) are two important aspects for modern Machine Learning systems. To enhance data privacy, recent machine learning models have been designed as a Federated Learning (FL) system. On top of that, additional privacy layers can be added, via Differential Privacy (DP). On the other hand, to improve explainability, ML must consider more interpretable approaches with reduced number of features and less complex internal architecture. In this context, this paper aims to achieve a machine learning (ML) model that combines enhanced data privacy with explainability. So, we propose a FL solution, called Federated EXplainable Trees with Differential Privacy (FEXT-DP), that: (i) is based on Decision Trees, since they are lightweight and have superior explainability than neural networks-based FL systems; (ii) provides additional layer of data privacy protection applying Differential Privacy (DP) to the Tree-Based model. However, there is a side effect adding DP: it harms the explainability of the system. So, this paper also presents the impact of DP protection on the explainability of the ML model, analyzing the obtained results for  SHAP (SHapley Additive exPlanations) and Mean Decrease in Impurity (MDI).

\end{abstract}


\section{Introduction}
Modern Machine Learning (ML) systems must follow compliance and legislation rules to support high levels of Data Privacy \cite{wei2020federated}. In parallel, there is a great concern from the industry to deploy ML models with better explainability \cite{zhang2022explainable}.

The term eXplainable Artificial Intelligence (XAI) refers to the capacity of the ML model to produce trackable outputs, implying in MLs with reduced number of features and less complexity. A promising approach for explainable ML is Decision Tree (DT) \cite{mienye2024survey}. A DT-based model has simpler internal structure than other highly-used models like Neural Networks or Support Vector Machine and still can have similar or better performance, mainly for tabular data \cite{grinsztajn2022tree} \cite{10305887}. Besides, the output of a DT can be trackable, since each decision node in the DT has a known set of rules.

Regarding data privacy, ML models have been driven to distributed approaches as Federated Learning (FL). In FL systems, the users' data is not transferred to a central training point. Instead, FL parameters are exchanged during the training phase, and the users' data is kept under the domain of the producer. However, the FL system still has open issues related to data privacy. This is because malicious agents can obtain the users' private data by processing the shared FL parameters. For instance, a malicious FL server can run a Gradient Inversion \cite{rodriguez2023survey} or a Membership Attack \cite{truex2019demystifying} to obtain sensitive data.

In order to achieve both, data privacy and explainability, this paper proposes a FL solution, called Federated EXplainable Trees with Differential Privacy (FEXT-DP), that: (i) is based on Decision Trees, since they are lightweight and have superior explainability than neural networks-based FL systems; (ii) provides additional layer of data privacy protection applying Differential Privacy (DP) to the Tree-Based model. It is important to notice these two goals are inter-related, since adding DP damages the explainability of the model. So, the contributions of this paper are:

\begin{itemize}
    \item To propose a Federated Learning model that jointly achieves explainability and extra-level of data privacy.
    \item To study how Differential Privacy impacts explainability on a Tree-based FL model.
\end{itemize}

This paper is organized as follows: Section \ref{sec:RW} introduces related work. Section \ref{sec:solution} details the proposed solution. Section \ref{sec:evaluation} describes the evaluation environment and the obtained results. Section \ref{sec:conclusion} presents conclusions and future work.

\section{Related Work}
\label{sec:RW}
Several studies, seeking eXplainable Artificial Intelligence (XAI), have integrated tree-based models into federated learning following different strategies. Li et al. \cite{li2023fedtree} developed FedTree, a system that incorporates Gradient Boosting Decision Trees (GBDT) into a federated setting. Their method employs a sequential, ensemble-based training process where subsequent models address the errors of prior ones. Although the approach includes security enhancements, its dependence on Homomorphic Encryption introduces a significant performance overhead.

In a different approach, Shen et al. \cite{shen2022efficient} design a framework for deploying random forests in distributed environments. The authors implement specific data partitioning techniques to better secure sensitive information from network sensors. A shortcoming of this distributed solution is the absence of a mechanism description for aggregating trees on a central FL server.

Addressing similar data privacy concerns, Souza et al. \cite{de2020dfedforest} introduce a distributed random forest approach that utilizes blockchain technology to record references to local models. This solution is able to preserve the global model's accuracy against potential sabotage by malicious participants. Separately, Zhao et al. \cite{zhao2024communication} proposed FL-DT, a method that bases its node-splitting criteria on the partial information at each decision tree node. This technique aims to estimate the global Gini Index bounds for every feature by gathering local statistics. A primary drawback is that the sequential exchange of this data demands an excessive number of communication rounds.

Other works focus on enhancing security protocols. Liu et al. \cite{liu2020federated} introduced a framework that offloads the task of identifying the best feature for each tree node to the clients, with the server determining the optimal split. Similarly, Feng et al. \cite{feng2019securegbm} created SecureGBM, a secured multi-party boosting system built as an extension of LightGBM that uses stochastic approximation to lower communication overhead. A weak point in SecureGBM is that the use of anonymous features reduces the model's explainability. In response, they propose the FED-EINI framework, which aims to restore explainability by assigning meaning to the features.

Maddock et al. \cite{maddock2022federated} propose a framework for the differentially private training of Gradient Boosted Decision Tree, for instance XGBoost, in the federated learning architecture. Similarly, Qin et al. \cite{qin2023differential} introduce a novel approach for allocating a privacy budget in a decision tree ensemble model. The core of this strategy is to link the allocation of the privacy budget directly to the depth of a leaf node. This method ensures a more granular and efficient use of the privacy budget, allowing for more precise control over the trade-off between privacy and model utility. However, different to Maddock et al. \cite{maddock2022federated} and  Qin et al. \cite{qin2023differential}, FEXT-DP focus on bagging-based decision tree approach.

Liu et al. \cite{liu2018differentially} present a solution to train private decision trees and to ensemble using differential privacy. The process involves two key procedures. First, internal nodes are selected based on a noisy maximal vote to preserve privacy. Additionally, a budget allocation strategy is designed to add less noise at larger depths, striking a balance between the true data and the added noise. Second, at the leaf nodes, the votes for each class are masked with Laplacian noise. By implementing these two procedures, the authors ensure that the entire process of growing a private decision tree is fully compliant with differential privacy principles. Despite the valuable findings highlighted by Liu et al. \cite{liu2018differentially}, these authors do not consider a federated learning architecture.

\section{Federated EXplainable Trees with Differential Privacy (FEXT-DP)}
\label{sec:solution}

\begin{figure}[h]
\centering
\includegraphics[width=8cm]{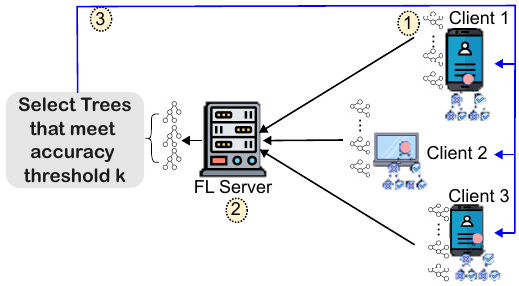}
\caption{Overview.}
\label{fig:overview}
\end{figure}

As can be observed on Fig. \ref{fig:overview}, the proposed approach, named Federated EXplainable Trees with Differential Privacy (FEXT-DP), is based on Decision Trees and are designed for Federated Learning systems. During the training phase, FEXT-DP performs three stages, which are described as follows: \textcircled{\raisebox{-0.9pt}{1}} the Federated Learning (FL) Clients train the tree-based model using its local dataset and send it to the server; \textcircled{\raisebox{-0.9pt}{2}} The FL server receives the tree-based models and aggregates them by selecting only trees that meet the minimum level of accuracy, called threshold $K$. The trees that do not meet this minimum are discarded;  \textcircled{\raisebox{-0.9pt}{3}} The FL server sends the new set of trees to the clients. After these three stages one training round is completed. The training rounds continue until the stopping criteria is reached.

\begin{figure}[h]
\centering
\includegraphics[width=8cm]{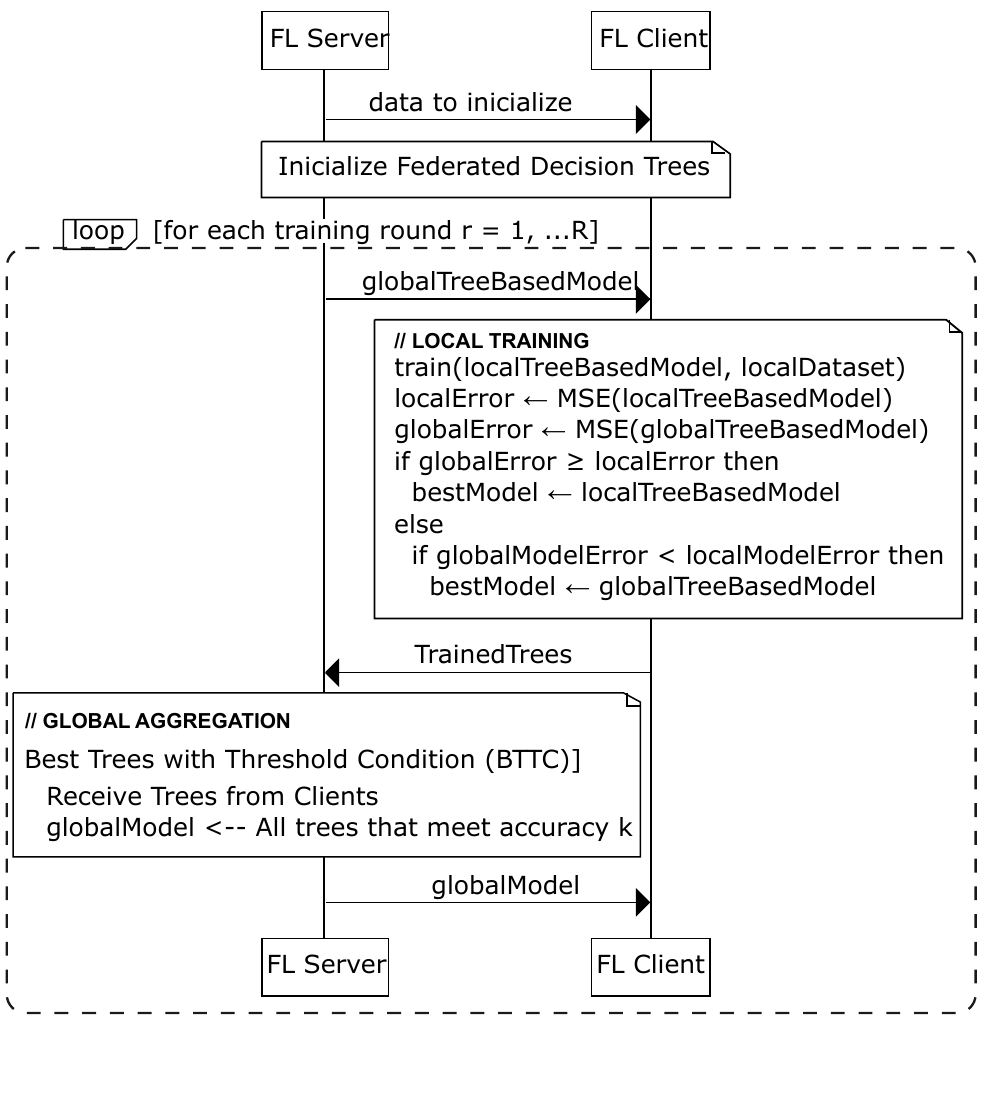}
\caption{Diagram: Detailed interaction between FL server and clients.}
\label{fig:diagram1}
\end{figure}

Fig. \ref{fig:diagram1} presents a diagram that further details the training interactions between FL Server and Clients. It is possible to notice that the training rounds start after the initial exchange of data. Then, in a particular round, after the client performs training, the client can compare if the model provided by the FL Server, i.e. global model, has lower Mean Square Error (MSE) than the most recent local model. The FL Client chose the set of trees with lower MSE. 

In the FL Server side, all the received trees are tested to verify their levels of accuracy. The method used in this work considers a minimum threshold K to be met by the trees. This procedure eliminates low performance trees from potential malicious FL Clients who may want to poison the FL model.


Algorithm \ref{algo:1} present how a FL Client train a Decision Tree with its local dataset. The aim of Algorithm \ref{algo:1} is not to present the whole training pseudocode, but to highlight the changes proposed to enable differential privacy in the Decision Trees. So, as can be observed, this algorithm introduces a differentially private method for finding the optimal split in a decision tree's training process. Instead of deterministically selecting the split with the highest information gain, it introduces a layer of privacy by using the exponential mechanism that considers the differential privacy budget, $\epsilon$. Similarly to a traditional Decision Tree training, the algorithm first calculates the information gain for every possible feature-threshold split. Then, it uses the exponential mechanism to assign a probability to each potential split, with higher probabilities given to splits that yield greater information gain. This process is a key part of differential privacy, as it ensures that the selected split is not directly tied to any single data point. The selection is then made randomly, with a ``roulette wheel" selection based on the calculated probabilities, returning a differentially private best split for the decision tree.



\begin{algorithm}
\footnotesize
\caption{best\_split\_with\_DP(X, y, $\epsilon$)}
\label{algo:1}
records $\gets [~]$\;
weights $\gets [~]$\;
probabilities $\gets [~]$\;

\For{each feature in $X$}{
    thresholds $\gets$ unique\_values(feature)\;

    \For{each threshold in thresholds}{
        gain $\gets$ information\_gain(feature, $y$, threshold)\;
        records.append((threshold, gain, feature\_index))\;
    }

    \For{each r in records}{
        gain $\gets$ records.getGain(r)\;
        weights.append\big($\exp(\frac{\epsilon \cdot gain}{2 \cdot sensibilidade})$\big)\;
    }

    \For{each weight in weights}{
        probabilities.append(weight / $\sum$(weights))\;
    }
}

record $\gets$ roulette\_wheel\_selection(records, probabilities)\;
threshold $\gets$ record.getThreshold()\;
feature\_index $\gets$ record.getFeature()\;

\Return{feature\_index, threshold}
\end{algorithm}

According to Algorithm \ref{algo:1}, differential privacy budget, represented by $\epsilon$, is applied to the information gain (See Line 11). Thus, the differential privacy budget, i.e., $\epsilon$, quantifies the maximum amount of privacy the algorithm can spend when releasing information about the information gain during the training. The DP budget is a limit on the total privacy loss. A smaller DP budget value means a stricter privacy guarantee, as it allows less information to be revealed.

A secondary effect of this use of $\epsilon$ in line 11 is when a malicious agent has access to a FEXT-DP tree. The malicious agent will not be able to retrieve the information gain with full accuracy and this represents a preventive measure against Membership Attacks \cite{truex2019demystifying}.

\section{Evaluation Performance}
\label{sec:evaluation}
In this section, we describe the conducted evaluation, present the obtained results for (i) FEXT-DP, (ii) federated decision tree with no differential privacy, and (iii) FedAVG, which is a traditional federated learning approach based on Neural Networks. 

This section is organized as follows: Section \ref{susec:dataset} describes the dataset used to perform this evaluation; Section \ref{subsec:implemetation} introduces the settings, implementation details and the selected metrics to measure the performance; Section \ref{subsec:results1} and \ref{subsec:results2} present and discuss the obtained results.

\subsection{Dataset}
\label{susec:dataset}
The proposed solution is designed for horizontal federated learning scenarios, where every client has the same set of features. For this evaluation, we consider a federated learning scenario where sensors are deployed in a residence to produce data for the local dataset. For each residence, there is a particular node that acts as a FL client, having access to the local data. 

To implement this scenario, we used the Appliance Energy Prediction Data (AEPD) provided by Candanedo et al. \cite{candanedo2017data}. AEPD holds information about the amount of spent energy in homes. AEPD contains 19,735 entries, providing the real energy consumed by residencies in Stambruges (Belgium), considering the period between January to May, 2016. Using AEPD, the machine learning model is trained taking as input: (i) temperature and humidity in various rooms, (ii) residence location, (iii) date, (iv) time (hh/mm), and some other features. The models must provide a daily prediction for the energy consumed by the residence, which is a regression problem for machine learning.


We split the data into training and testing sets, allocating 80\% to training and 20\% to testing. To imitate a distributed environment, the training data was divided into 20 chunks, each chunk was generated from a general seed, which all clients have access to, added to the client id, to make each sample unique.

\subsection{Implementation and Performance Metrics}
\label{subsec:implemetation}



Federated EXplainable Trees with Differential Privacy
(FEXT-DP) has been implemented in Python,
using the following libraries and versions: pandas 2.2.3, scikit-learn 1.7.dev1+dp, numpy 2.2.5, and scipy 1.15.2. This version of scikit-learn is based on version 1.7.0, but has modifications necessary to implement differential privacy.

To perform this evaluation, we consider twenty FL clients and one server, all of them running in the PC with the following settings: Operating System: Arch Linux x86\_64, with kernel 6.12.43-1-lts, 64 bits, Memory RAM: 12288 MiB, Intel Core i5-4590 CPU 3.30GHz x 4, Intel HD Graphics 4600 / AMD Caicos and disk capacity of 500,10 GB.


For the FEXT-DP threshold K, it was defined as 0.5. Regarding FedAVG, it is implemented as a Neural Network with six layers, in total 600 neurons, using ReLU as activation function, batch normalization on second and four layers, and backpropagation. In every round, FedAVG runs a maximum of 40 epochs.


The following performance metrics have been considered for this evaluation:
\begin{itemize}
    \item \textbf{Pearson Correlation:} The Pearson correlation method is the most common method to use for numerical variables; It is a similarity-based approach that compares one data object with another, attribute by attribute, usually summing the squares of the differences in magnitude for each attribute, and using the calculation to compute a final outcome, known as the correlation score \cite{BERMAN2016135}. The final outcome is a value between - 1 and 1, where 0 is no correlation, 1 is total positive correlation, and - 1 is total negative correlation \cite{NETTLETON201479}. 

    \item \textbf{MSE:} This is a metric that evaluates a model's performance by calculating the average of the squared differences between predicted and actual values.

    \item \textbf{Mean Decrease in Impurity (MDI):} This metric is used to measure the explainability of the models. It measures how much each feature contributes to reducing impurity (e.g., Gini impurity or entropy) in tree-based models. So, high values of MDI for a feature means that it has more importance for the performance of the model.
\end{itemize}

\subsection{Obtained Results: The impact of Differential Privacy on the Training Performance}
\label{subsec:results1}
\begin{figure}[h]
     \centering
     \begin{subfigure}[h!]{0.5\textwidth}
         \centering
            \includegraphics[width=0.9\linewidth]{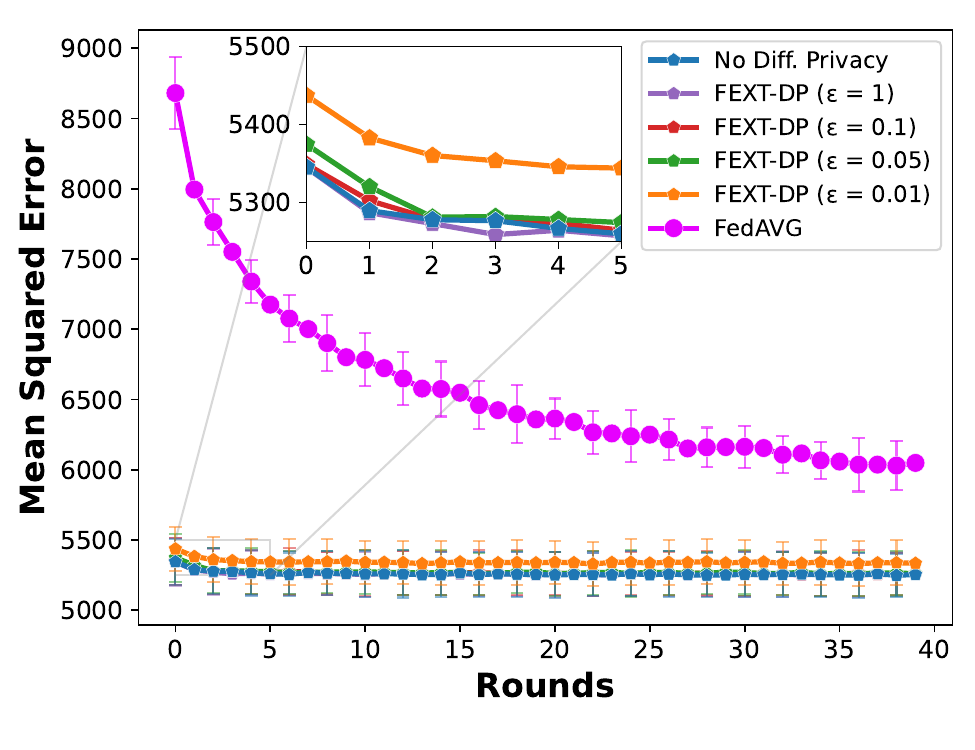}
         \caption{Mean Square Error (MSE).}
         \label{fig:result1}
     \end{subfigure}
     
     \begin{subfigure}[h!]{0.5\textwidth}
         \centering
            \includegraphics[width=0.9\linewidth]{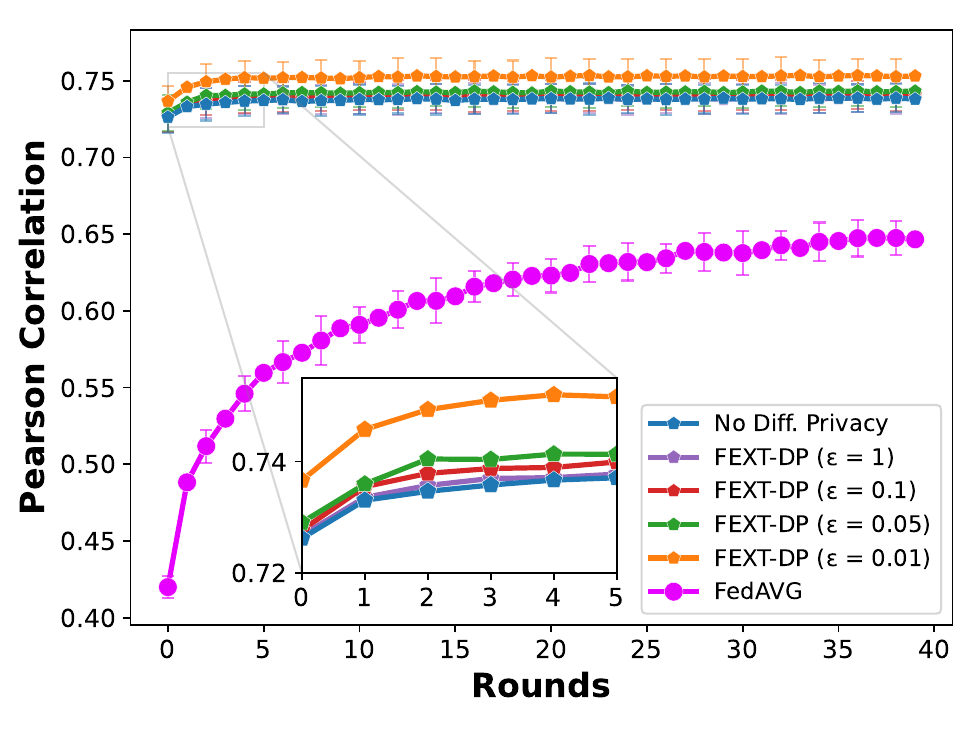}
         \caption{Pearson Correlation.}
         \label{fig:result2}
     \end{subfigure}
\caption{The impact of Differential Privacy on Performance Over the Training Rounds.}
\end{figure}

As can be observed in Fig. \ref{fig:result1}, the obtained MSE for FEXT-DP are similar for all epsilon variations, except for the lowest differential privacy budget, i.e., $\epsilon =0.01$. With a $\epsilon =0.01$, the provided privacy reaches the highest value in this evaluated setting. So, it is expected to have less MSE performance, due to the DP mechanism inserted in line 11 of the algorithm \ref{algo:1}. However, it is worth noting that FEXT-DP with $\epsilon=0.01$ is 1.18\% MSE higher than decision trees with no differential privacy.

Regarding the other FEXT-DP variants and the approach with no DP, the MSE values obtained are around 5500 in round 1. These values decrease to around 5250 after round 4. In comparison, FedAVG begins with a higher MSE value (8568), which means that FEXT-DP begins with  35\% less MSE than FedAVG. It is possible to notice, FedAVG over the 40 rounds decreases the MSE, reaching 6000, which is a worse performance than FEXT-DP.

Fig. \ref{fig:result2} presents the obtained results for FEXT-DP, FedAVG, traditional Decision Trees with no differential privacy in terms of Pearson Correlation. For Pearson Correlation, results near to 1 means better correlation. A similar behavior in the performance is observed compared with MSE. FedAVG begins the training rounds with significantly worse performance than FEXT-DP, but it presents an improvement curve. Regarding FEXT-DP, the approach with the highest DP budget ($\epsilon = 0.01$) presents the worst Pearson Correlation performance.

\subsection{Obtained Results: The impact of Differential Privacy on the Explainability}
\label{subsec:results2}
\begin{figure}[b!]
\centering
\includegraphics[width=8.5cm]{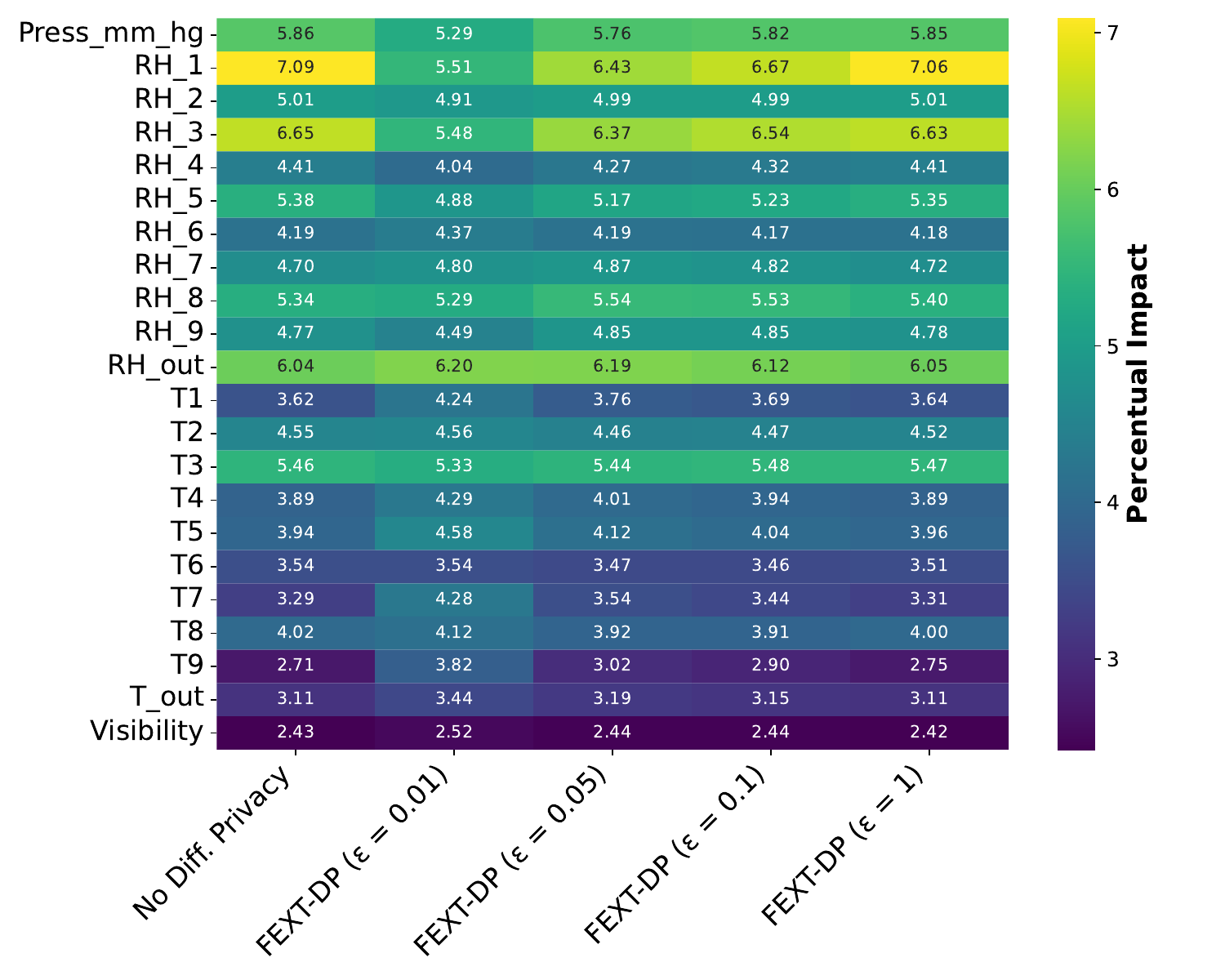}
\caption{Heatmap of Mean Decrease in Impurity (MDI).}
\label{fig:heatmap}
\end{figure}

\begin{figure}[b!]
\centering
\includegraphics[width=0.75\linewidth]{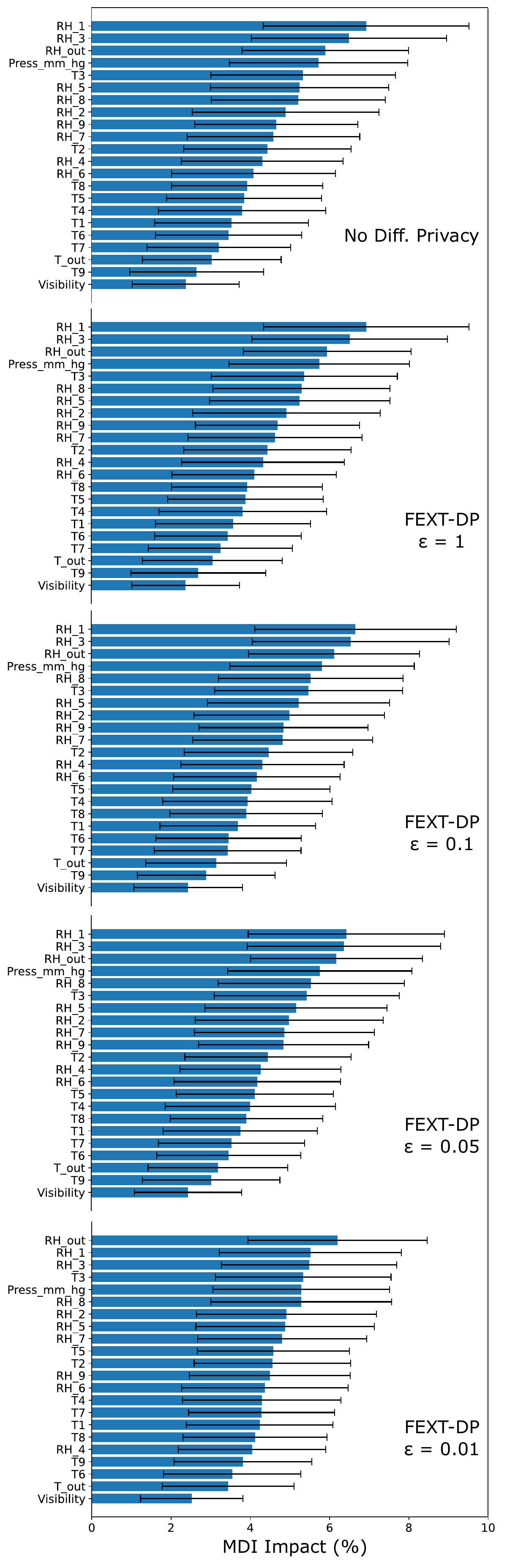}
\caption{The impact of Differential Privacy on Explainability Based on the Mean Decrease in Impurity (MDI).}
\label{fig:MDI}
\end{figure}

Mean Decrease in Impurity (MDI) is the metric used in this work for explainability of the Decision Tree models. The obtained MDI values can be observed in Fig. \ref{fig:MDI} and \ref{fig:heatmap}. For this metric, each feature of the model has an index (\%), showing which features have more impact for the model's performance. To better understand the meaning of the features in Figure \ref{fig:MDI} and Figure \ref{fig:heatmap}, refer to Table \ref{tab:features}.

{\setlength{\tabcolsep}{2pt}
\begin{table}[h]
    \centering
\begin{tabular}{ll}

\hline
\hline
\textbf{Features} & \textbf{Meaning}\\
\hline
RH\_1 to RH\_9 & It refers to Room Humidity in nine rooms of the residence. \\
\hline
RH\_out & It is the measured Humidity outside the residence.\\
\hline
Press\_mm\_hg & It refers to the environment pressure inside the residence.\\
\hline
T1 to T9 & It is the Temperature in nine rooms of the residence.\\
\hline
T\_out & It is the measured temperature outside the residence.\\
\hline
Visibility & This is the visibility measured in a nearby weather station. \\
\hline
\hline
\end{tabular}
\caption{Description of the Most Relevant Features for FEXT-DP.}
\label{tab:features}
\end{table}
}

As can be observed in Fig. \ref{fig:MDI}, the feature ``RH\_1" is the most relevant feature for FEXT-DP, except for FEXT-DP with $\epsilon=0.01$. Besides, the ``visibility" feature received less importance for all tested approaches. Moreover, it can noticed how the MDI is reduced comparing the approach with no differential privacy with the four variants of FEXT-DP. It means that applying differential privacy on the methods causes significant changes in the order of most important features for all tree-based models.

Fig. \ref{fig:heatmap} presents the MDI values in a heatmap in order to better visualize how these results change among the evaluated approaches. Dark colors represent low values of MDI, while light colors mean high values.

In Fig. \ref{fig:heatmap}, comparing the first and second column, i.e., No diff. Privacy vs FEXT-DP ($\epsilon=0.01$), it is possible to notice a significant difference. More specifically, FEXT-DP with $\epsilon=0.01$ has lower MDI values for all the features. It means that the applied differential privacy impacts the explainability of the model by spreading the relevance over all features. In other words, there is a lower prevalence of a small group of features, which affects the explainability because the model's output becomes trackable.

\begin{figure}[h]
    \centering
    \begin{subfigure}{0.5\textwidth}
        \centering
        \includegraphics[width=1\linewidth]{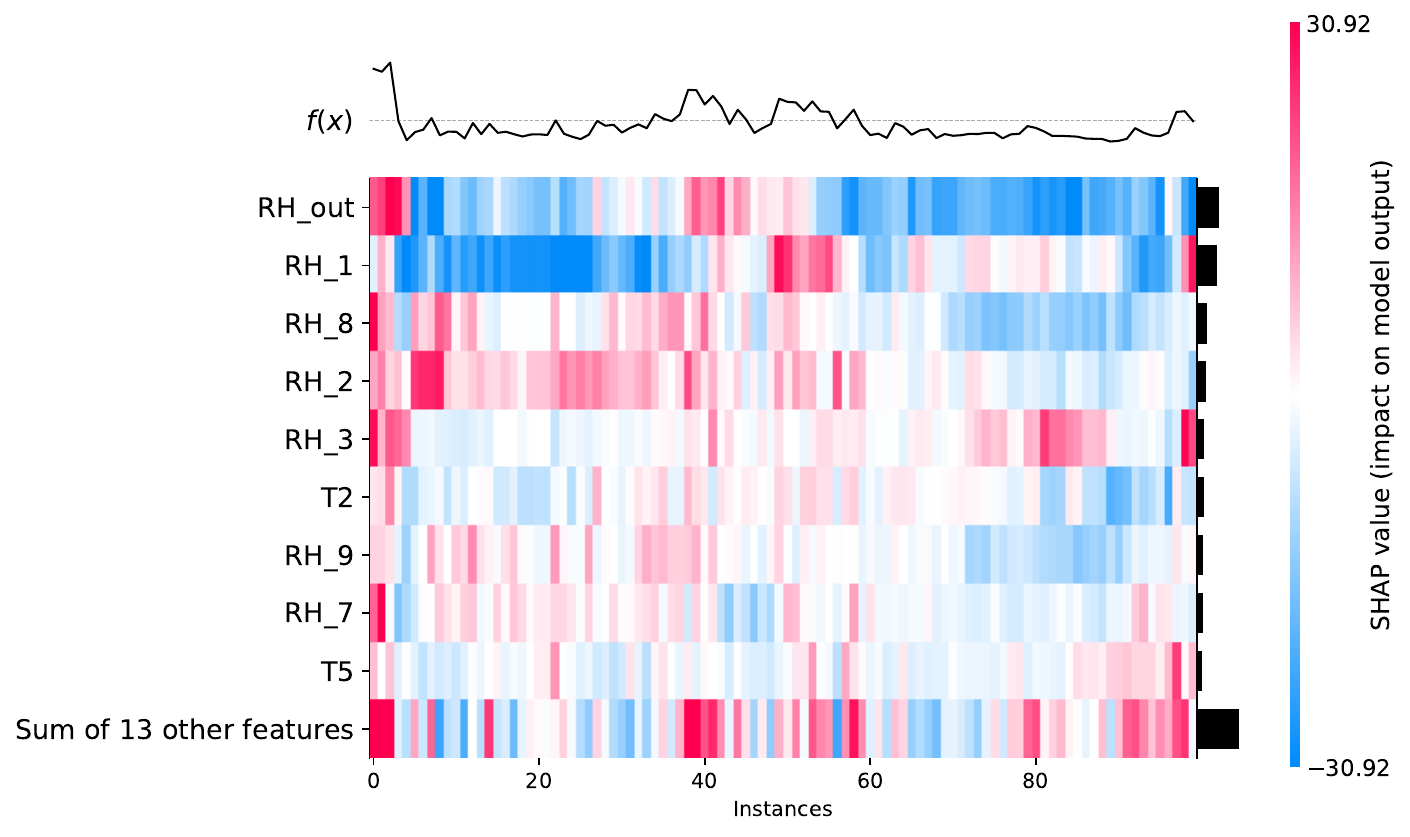}
        \caption{$\epsilon = 0.01$ - Highest level of privacy. }
    \end{subfigure}%
    
    \begin{subfigure}{0.5\textwidth}
        \centering
        \includegraphics[width=1\linewidth]{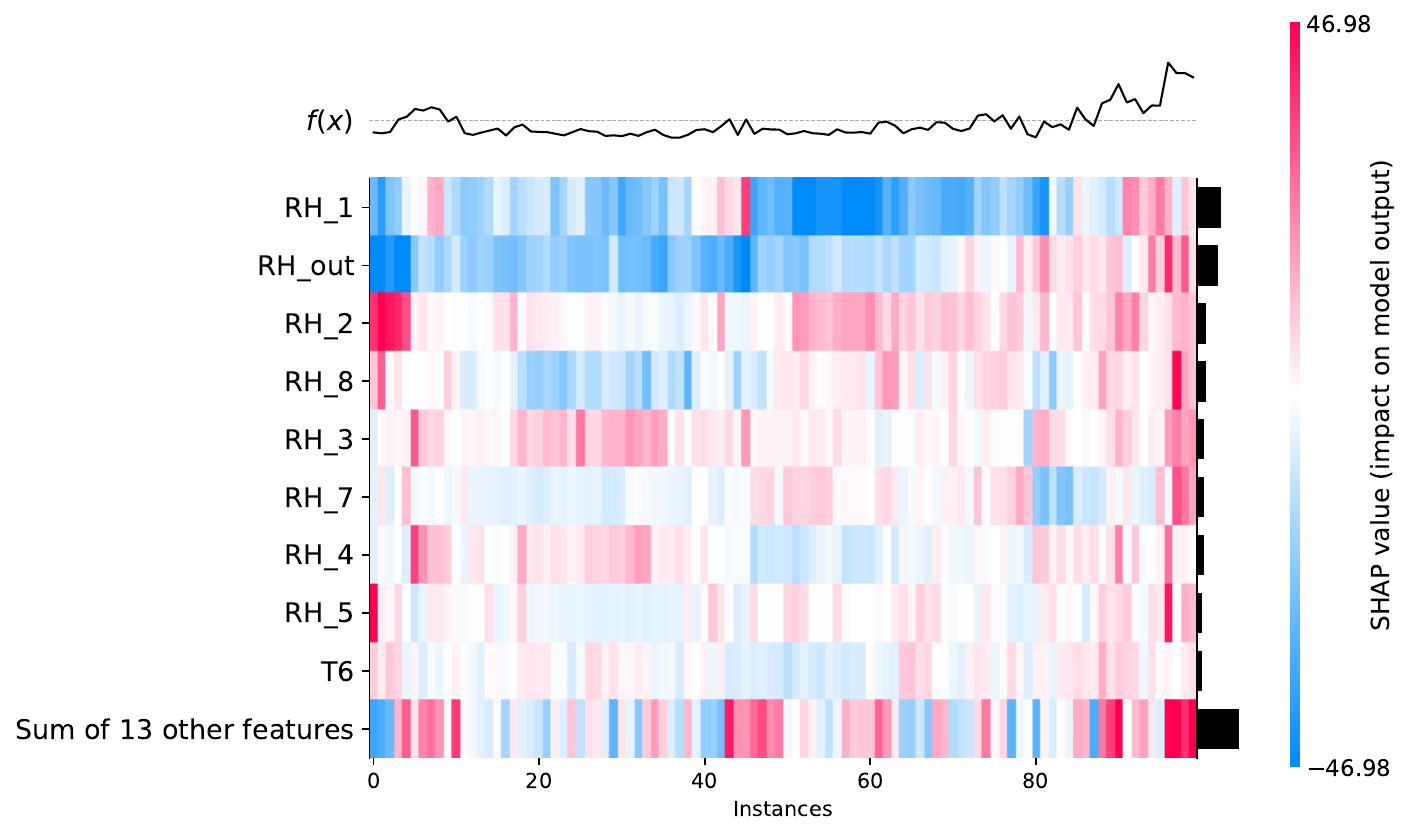}
        \caption{No differential privacy.}
    \end{subfigure}
    \caption{Heatmap of obtained SHAP values over the instances.}
    \label{SHAP-heatmap}
\end{figure}

\begin{figure}[h]
    \centering
    \begin{subfigure}[t]{0.45\textwidth}
        \centering
        \includegraphics[width=1\linewidth]{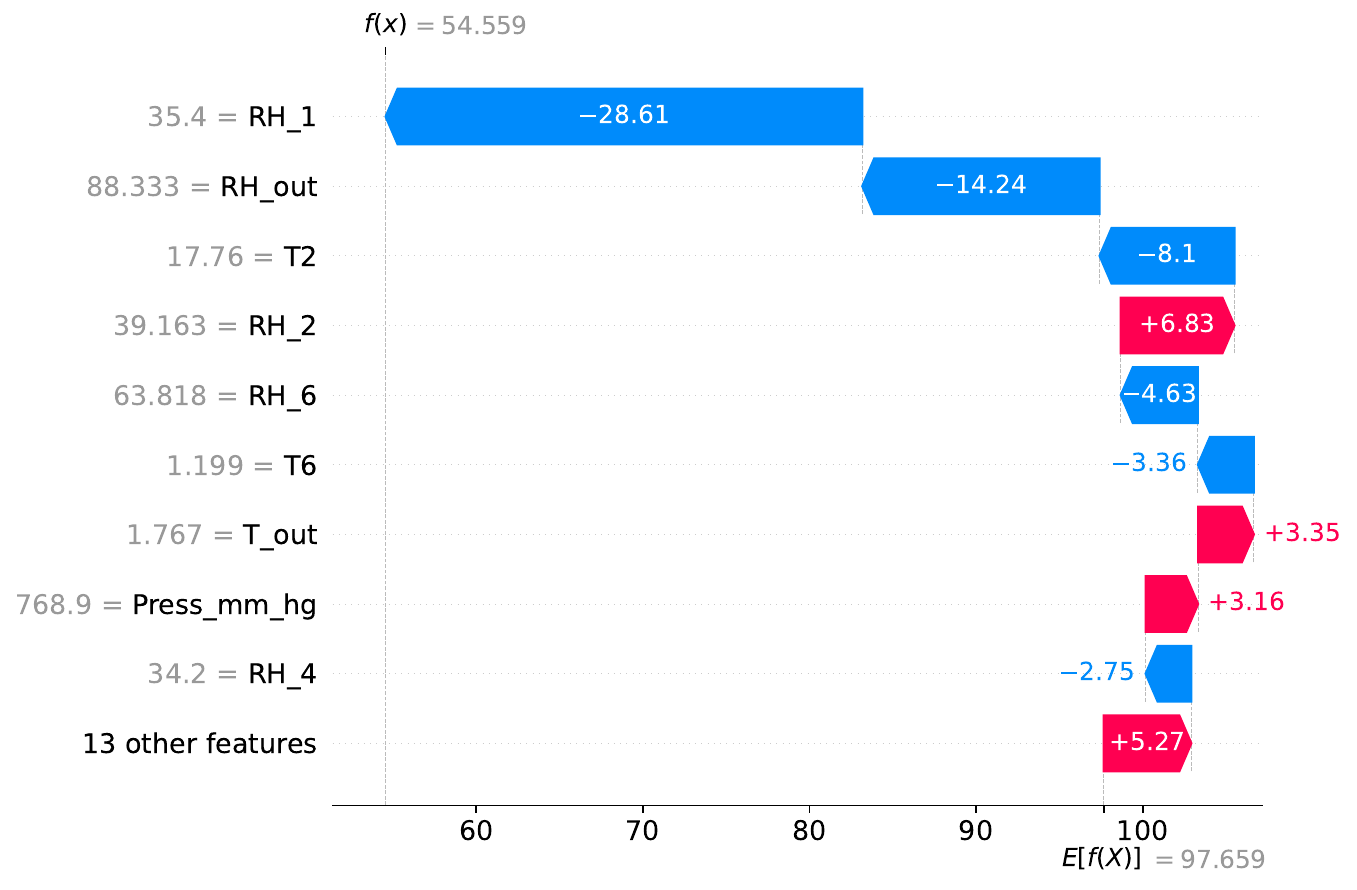}
        \caption{$\epsilon = 0.01$ - Highest level of privacy. }
    \end{subfigure}%
    
    \begin{subfigure}[t]{0.45\textwidth}
        \centering
        \includegraphics[width=1\linewidth]{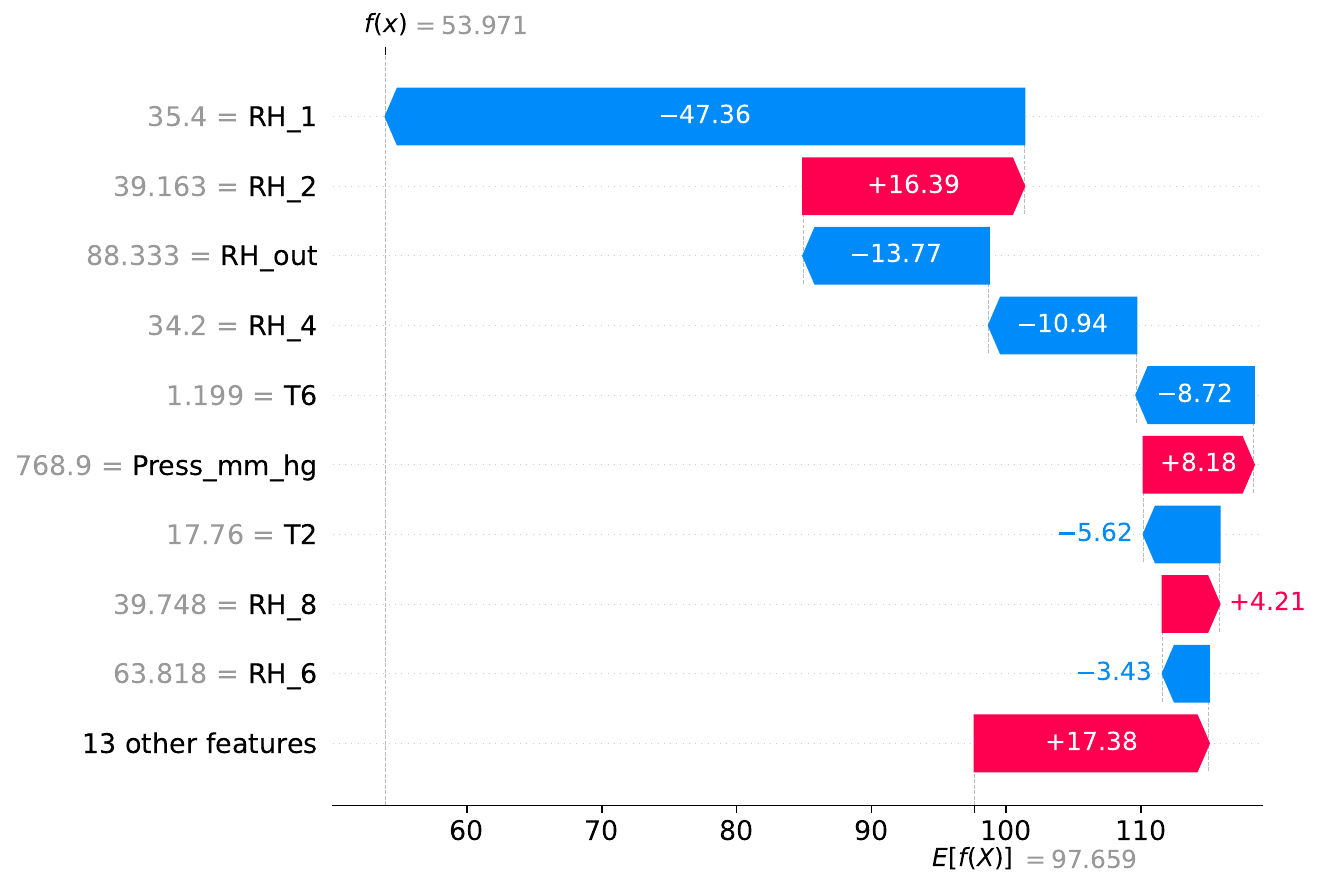}
        \caption{No differential privacy.}
    \end{subfigure}
    \caption{Waterfall chart based on the obtained SHAP values for most relevant features.}
    \label{SHAP-waterfall}
\end{figure}

\begin{figure*}[h]
    \centering
    \begin{subfigure}[t]{0.5\textwidth}
        \centering
        \includegraphics[width=1\linewidth]{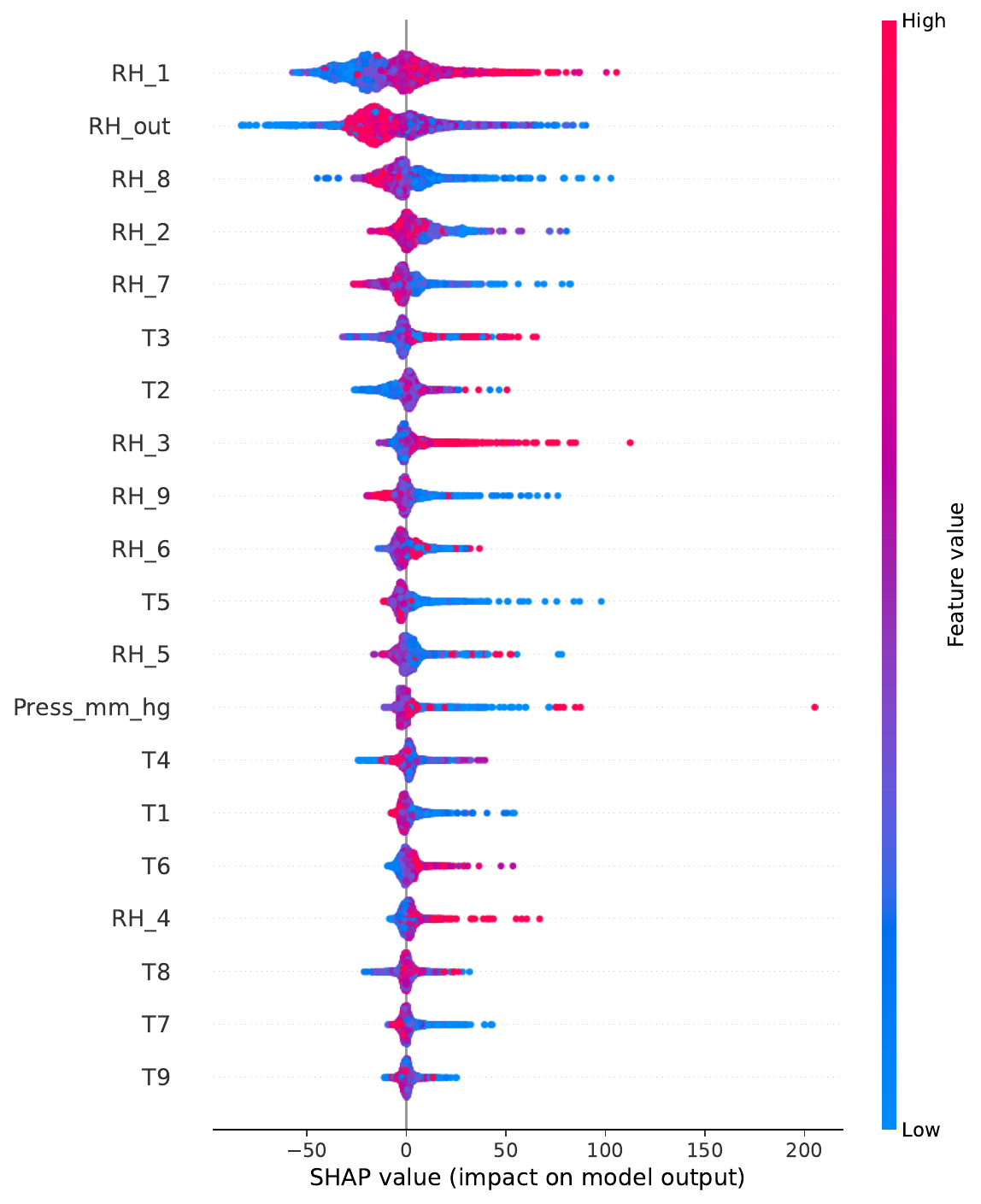}
        \caption{$\epsilon = 0.01$ - Highest level of privacy. }
    \end{subfigure}%
    ~ 
    \begin{subfigure}[t]{0.5\textwidth}
        \centering
        \includegraphics[width=1\linewidth]{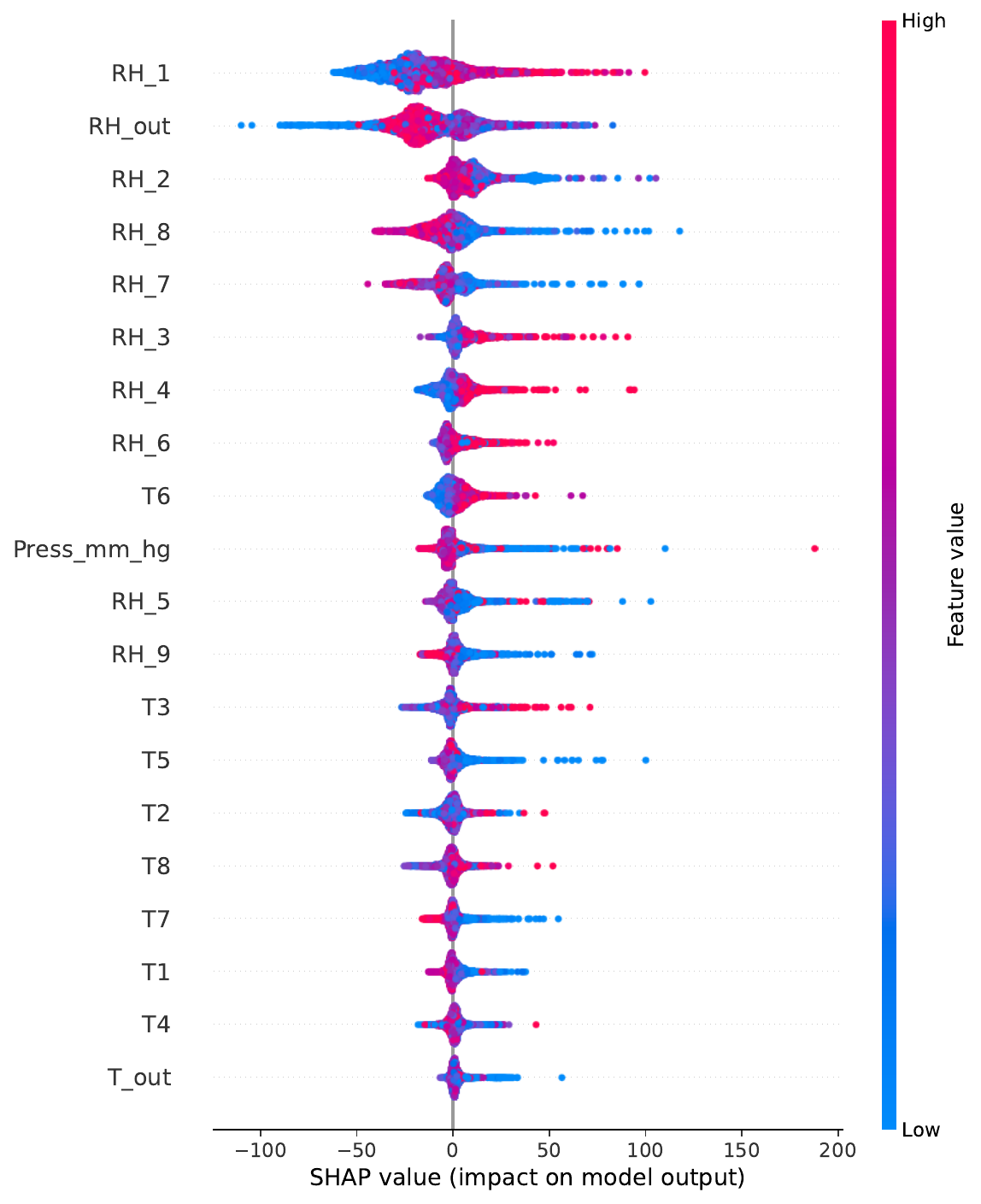}
        \caption{No differential privacy.}
    \end{subfigure}
    \caption{Obtained SHAP values for each feature.}
        \label{SHAP-points}
\end{figure*}

Fig. \ref{SHAP-heatmap}, \ref{SHAP-waterfall} and \ref{SHAP-points} present the obtained SHAP (SHapley Additive exPlanations), which quantifies how important each input variable is to a model for producing an outcome. Fig. \ref{SHAP-heatmap} depicts the dataset instances on the x-axis, the model inputs on the y-axis, and the SHAP values encoded on a color scale. An instance is a sample from the dataset being analyzed. It is possible to notice the model's output above the heatmap and a bar plot on the right-hand side illustrates the global importance of each input feature.

Fig. \ref{SHAP-waterfall} explains the underlying contributions of each feature to a single prediction, in this case it is the base value 97.659. As can be observed, the waterfall chart shows at the bottom the model's expected value ($E[f(X)]$). Each subsequent row illustrates the additive contribution of individual features, red for positive and blue for negative impacts, transitioning the value from the background expectation to the final model prediction. Comparing Fig. \ref{SHAP-waterfall}.a and Fig. \ref{SHAP-waterfall}.b, it is observable that with $\epsilon=0.01$ there is a different order of feature contributions for the final outcome. For instance, the feature $t_{out}$ appears in Fig \ref{SHAP-waterfall}.a and it is listed in the \textit{13 other features} in Fig \ref{SHAP-waterfall}.b, which are the ones with less contributions.

Instead of presenting a single prediction, Fig. \ref{SHAP-points} presents a beeswarm plot with the SHAP values. This plot offers a global distribution of feature effects. It visualizes the density of SHAP values for every feature across all predictions, allowing us to identify non-linear relationships and the overall variance in feature contribution. Every dataset sample is represented by a single dot on each feature row and the distribution of points reveals how a feature influences the model's prediction. For instance,  Fig. \ref{SHAP-points}.a reveals that low values for RH\_1 contribute to producing negative SHAP values.

Generally, a differential privacy budget of $\epsilon = 0.01$ introduces only a minor perturbation to the explainability of the tree-based model. While SHAP values reveal shifts among the most significant features, the internal feature operations remain largely consistent when comparing the non-private baseline to the FEXT-DP ($\epsilon = 0.01$) implementation.

\section{Conclusions and Future Works}
\label{sec:conclusion}
Machine Learning (ML) systems today need to adhere to compliance and legislation rules that ensure a high level of data privacy. In addition, there is a significant effort from the industry to deploy ML models with greater explainability, seeking eXplainable Artificial Intelligence (XAI). Explainability refers to a model's ability to produce clear, traceable outputs. This often means using models with a reduced number of features and a less complex structure. 

This paper proposes Federated EXplainable Trees with Differential Privacy (FEXT-DP), which is a Federated Learning model based on Decision Trees with additional layer of privacy provided by Differential Privacy. The decision trees used in FEXT-DP are ideal for this task because they are both lightweight and inherently more explainable than neural network-based FL systems. To enforce privacy preserving, the solution applies Differential Privacy (DP) directly to the tree-based model. 

Despite the significant advances, there are some investigations remaining for future work. We plan to enhance our FEXT-DP by incorporating pruning algorithms and a client selection mechanism to improve its efficiency and overall performance. Our future research will also explore a novel design for FEXT-DP with aspects to be inserted to achieve more explainability of the decision trees. Beyond our current findings, we will analyze other critical performance metrics, including training latency, RAM memory usage, and network traffic, to better understand the system's real-world behavior and scalability.

\section*{Acknowledgement}
This research was funded in part by CNPq, under grant \#444978/2024-0. This study was also financed in part by the Coordenação de Aperfeiçoamento de Pessoal de Nível Superior - Brasil (CAPES) - Finance Code 001. 

\bibliographystyle{ieeetr}
\bibliography{paper}

\end{document}